\documentclass{article}
\usepackage{graphicx}
\usepackage{amsmath}
\usepackage{blindtext}

\usepackage[numbers]{natbib} 

\usepackage[final]{tackling_climate_workshop_style}

\usepackage{tackling_climate_workshop_style}

\usepackage[utf8]{inputenc} 
\usepackage[T1]{fontenc}    
\usepackage{hyperref}       
\usepackage{url}            
\usepackage{booktabs}       
\usepackage{amsfonts}       
\usepackage{nicefrac}       
\usepackage{microtype}      

\title{Facade Segmentation for Solar Photovoltaic Suitability}

\author{%
  Ayca Duran \\
  ETH Zurich \\
  Architecture and Building Systems \\
  Zurich, Switzerland \\
  \texttt{duran@arch.ethz.ch} \\
  \And
  Christoph Waibel \\
  VITO \\ 
  CRIB \\ 
  Mol, Belgium \\
  \texttt{christoph.waibel@vito.be} \\
  \And
  Bernd Bickel \\
  ETH Zurich \\
  Computational Design Lab \\
  Zurich, Switzerland \\
  \texttt{bickelb@ethz.ch} \\
  \And
  Iro Armeni \\
  Stanford University \\
  Gradient Spaces Lab \\
  Stanford, CA, US \\
  \texttt{iarmeni@stanford.edu} \\
  \And
  Arno Schlueter \\
  ETH Zurich \\
  Architecture and Building Systems \\
  Zurich, Switzerland \\
  \texttt{schlueter@arch.ethz.ch} \\
}

\begin{document}

\maketitle

\begin{abstract}
   Building integrated photovoltaic (BIPV) facades represent a promising pathway towards urban decarbonization, especially where roof areas are insufficient and ground‑mounted arrays are infeasible. Although machine learning-based approaches to support photovoltaic (PV) planning on rooftops are well researched, automated approaches for facades still remain scarce and oversimplified. This paper therefore presents a pipeline that integrates detailed information on the architectural composition of the facade to automatically identify suitable surfaces for PV application and estimate the solar energy potential. The pipeline fine-tunes SegFormer-B5 on the CMP Facades dataset and converts semantic predictions into facade-level PV suitability masks and PV panel layouts considering module sizes and clearances. Applied to a dataset of 373 facades with known dimensions from ten cities, the results show that installable BIPV potential is significantly lower than theoretical potential, thus providing valuable insights for reliable urban energy planning. With the growing availability of facade imagery, the proposed pipeline can be scaled to support BIPV planning in cities worldwide.
\

\end{abstract}

\section{Introduction}
The energy use of buildings during their operation accounts for nearly 28\% of global carbon emissions \cite{united_nations_environment_programme_2023_2024}. Building-integrated photovoltaics (BIPV) offer a promising pathway to reduce these emissions, with cost reductions and technological advances driving a wider adoption. Although rooftop photovoltaic (PV) design has been extensively researched, automated approaches to facade photovoltaic energy remain underexplored despite its relevance to urban renewable energy \cite{faes_building-integrated_2025, duran_review_2025}. Especially in high-rise and mixed-use developments, where roof space is limited and ground-mounted systems are constrained by land availability or regulations, BIPV can be a viable design choice. However, barriers to facade BIPV integration include current decision support workflows that are slow and tedious, expert-dependent, and difficult to scale, thus delaying a wider adoption despite the benefits \cite{faes_building-integrated_2025}.

A range of Geographic Information System (GIS)-enabled tools exist for PV potential assessment, which typically rely on rooftop geometries, satellite imagery, and irradiance databases, to identify installation areas and estimate energy yields. Some extend to facade applications, such as Sonnenfassade.ch \cite{portmann_sonnendachch_2019} and PVWatts \cite{dobos_pvwatts_2014}, but these often treat facades as uniform surfaces and overlook their complex architectural qualities. Although archetype-based approaches exist \cite{saretta_calculation_2020}, a thorough consideration of facade-level details and obstructions remains an understudied area. Recent work has applied conditional adversarial networks \cite{isola_image--image_2016} to segment facade imagery using the CMP Facades dataset \cite{hutchison_spatial_2013} and estimated the theoretical PV potential assuming that the raw wall share of a facade can be fully covered with BIPV panels \cite{duran_estimating_2023}. Although this approach automatically identifies PV-eligible wall surfaces, it does not progress to practical system design, such as feasible panel layouts.

In this study, we introduce a segmentation-and-layout workflow that bridges the gap between identifying eligible surfaces and a practical system design. This automated workflow includes (1) segmenting facade components, (2) constructing PV-suitability masks by excluding openings and protrusions with clearance buffers, and (3) generating practical panel layouts based on module size. This produces estimates in physical units (area in m², peak power in kWp, and annual energy yield in kWh) and layout visualizations suitable for integration into solar cadastres or retrofit planning.

\section{Data and Method}

\textbf{Datasets.} We use two datasets for model training and application. The CMP Facades dataset \cite{hutchison_spatial_2013} provides annotations per pixel for 13 classes: \textit{background, facade, window, door, cornice, sill, balcony, blind, molding, deco, pillar, shop} and \textit{unknown}. The data set consists of 606 annotated images. The \emph{unknown} class is ignored in both loss computation and metric evaluation. To demonstrate the conversion from pixel fractions to energy yield calculations, 373 facades are randomly sampled from the Large Scale Architectural Asset (LSAA) dataset \cite{zhu_large-scale_2022}. In the application phase, ortho-rectified images and location data provided in the LSAA dataset are utilized. The dimensions and orientation of the facade are manually obtained from Google Earth, a platform easily accessible to users.

\textbf{Model and implementation details.}  
We adapt the classifier head of SegFormer-B5 model \cite{xie_segformer_2021}, initialized from the ADE20K\cite{zhou_scene_2017}-finetuned checkpoint, to the 13 classes of CMP Facade dataset. The images are resized to $640\times640$ pixels. Data augmentation includes horizontal flips, small shifts, scaling and rotation, brightness/contrast adjustment, HSV jitter, and light Gaussian blur (Appendix~\ref{app:aug}). The model is trained using the AdamW optimizer (learning rate $2 \times 10^{-4}$, weight decay $10^{-2}$) with a cosine learning rate schedule and 5\% warm-up. The batch size is 4, constrained by GPU memory. The loss combines inverse-frequency class-weighted cross-entropy and $0.5\times$ soft Dice loss over non-\emph{unknown} classes. Early stopping (patience 10) is applied within 80 epochs. At inference, logits are bilinearly upsampled and argmax operation is applied per pixel. The hyperparameters are tuned following a small grid-search (Appendix~\ref{app:hyper}).

\textbf{PV mask construction.}  
PV-suitable masks are derived from semantic predictions by (i) selecting pixels classified as \emph{facade}, (ii) removing \emph{window}, \emph{door}, and \emph{shop} pixels with dilated buffers to account for frames and installation clearances, (iii) excluding \emph{balcony}, \emph{cornice}, \emph{sill}, \emph{molding}, \emph{deco}, \emph{pillar}, and \emph{blind} with class-specific buffers and area thresholds, and by (iv) removing small connected components below a minimum area threshold equivalent to one panel size with buffers.  

Let $P$ denote PV-suitable pixels and $F$ all facade pixels. The PV fraction per image is then:
\begin{equation}
f_{\mathrm{PV}} = \frac{|P|}{|F|}.
\end{equation}
With known facade area $A_{\mathrm{facade}}$ (m$^2$), the PV-suitable area is:
\begin{equation}
A_{\mathrm{PV}} = f_{\mathrm{PV}} \times A_{\mathrm{facade}} \quad (\mathrm{m}^2).
\end{equation}

\textbf{Panel layout generation and energy yield.}  \label{panel_layout_energy}
To translate PV-suitable areas into installable designs, we fit standard PV module grids inside the predicted mask. The largest axis-aligned rectangle fully contained within the mask is then located using a histogram-based maximal-rectangle search. Within this rectangle, we test both portrait and landscape orientations of common \textit{large (L)} and \textit{small (S)} module sizes with a fixed inter-panel gap (Appendix~\ref{app:panel_placement}). The number of modules along each axis is computed by the integer division of the rectangle size by the module-plus-gap footprint. The orientation that maximizes the count while minimizing leftover space is selected.

Given a rating for standard test conditions (STC) per module \(P_{\mathrm{module,STC}}\) and the fit module count \(N_{\mathrm{modules}}\), the installed peak DC capacity is
\begin{equation}
P_{\mathrm{DC,STC}} = P_{\mathrm{module,STC}} \times N_{\mathrm{modules}} \quad (\mathrm{kWp}),
\end{equation}

We also calculate the location- and orientation-specific annual energy yield ($E_{\mathrm{PV,ann}}$) using PVGIS24 \cite{huld_new_2012}. The details of the calculation can be found in the Appendix~\ref{app:pvgis}. The code is available on GitHub\footnote{\url{https://github.com/ycdrn/segment4pvlayout}}.

\section{Experiments and Results} \label{sec:results}
 
\textbf{Setup and evaluation metrics.}  
We use the CMP Facades dataset for training, validation, and testing \cite{hutchison_spatial_2013}.  
Segmentation performance for 12 classes is reported using dataset-level mean Intersection-over-Union (mIoU) and mean over per-image mIoU values, as well as pixel accuracy, macro precision, recall, and F1-score over all classes.
The formal definitions of these metrics are provided in Appendix~\ref{app:metrics}. Experiments ran on a Windows workstation (Ryzen~9~7950X; RTX~4090~24\,GB; driver~560.94/CUDA~12.6).

\textbf{Segmentation performance.}  
Our SegFormer-B5 outperforms classical Auto–Context baselines on CMP.  
Dataset-level mIoU is $0.52$ (per-image mean mIoU $0.44$), with macro precision/recall/F1 of $0.68/0.68/0.68$. To contextualize difficulty, we include two baselines: Uniform Random (each pixel sampled uniformly from the 12 evaluated classes) and Majority Class (all pixels predicted as background), evaluated with the same metrics. The IoU score per class is highest for large, distinct classes such as \emph{background} and \emph{window}, and lowest for small, diverse elements such as \emph{pillar} (Table~\ref{tab:cmp}). We attribute this gap to dataset class imbalance and resolution effects from the stride-4 decoder and bilinear upsampling, although we achieved performance improvements in smaller classes with data augmentation efforts.

\begin{table}[t]
\centering
\small
\begin{tabular}{lccccc}
\toprule
Method & mIoU & Pixel Accuracy & Macro Precision & Macro Recall & Macro F1 \\
\midrule
\textbf{Ours (SegFormer-B5)} & \textbf{0.52} & \textbf{0.76} & \textbf{0.68} & \textbf{0.68} & \textbf{0.68} \\
Auto--Context (ST3) \cite{gadde_efficient_2016} & 0.36 & 0.66 & \textemdash & 0.49 & \textemdash \\
Auto--Context (PW3) \cite{gadde_efficient_2016} & 0.38 & 0.68 & \textemdash & 0.49 & \textemdash \\
Baseline – Uniform Random & 0.03 & 0.08 & 0.08 & 0.06 & 0.05 \\
Baseline – Majority Class & 0.03 & 0.24 & 0.02 & 0.08 & 0.03 \\
\bottomrule
\end{tabular}
\caption{Segmentation performance on the CMP Facades. (–: unavailable metric in the cited study.)}
\label{tab:metrics_summary}
\end{table}

\begin{table}[t]
\centering
\small
\begin{tabular}{lcccccccccccc}
\toprule
Method & Bg & Fac & Win & Door & Corn & Sill & Balc & Blind & Mold & Deco & Pill & Shop \\
\midrule
Ours  & 0.74 & 0.62 & 0.69 & 0.46 & 0.51 & 0.48 & 0.50 & 0.56 & 0.37 & 0.52 & 0.33 & 0.43 \\
\bottomrule
\end{tabular}
\caption{Per-class IoU on CMP Facades dataset.}
\label{tab:cmp}
\end{table}

\textbf{Class-share on the facade surface.}
PV suitability of a facade is commonly reported as a fraction of facade area \citep{duran_estimating_2023}. Therefore, we additionally present a domain-specific metric that compares the predicted and true shares of three class groups relative to facade surface: \emph{wall}, glazing (\emph{window}+\emph{shop}), and others. 
The predicted pixel share for major facade groups closely matches the ground truth (GT):  
\emph{wall} $47.42\%$ (GT: $49.39\%$), \emph{windows+shop} $16.48\%$ (GT: $16.08\%$), other classes $36.10\%$ (GT: $34.53\%$).  
The wall share error is $-1.97$ percentage points (pp). Across individual images, the mean absolute error is $7.29$ pp and the root mean square error is $12.17$ pp.

\textbf{Application to LSAA.}  
The trained pipeline is applied to 373 facades from the LSAA dataset with known facade areas. For each facade, PV fraction $f_{\mathrm{PV}}$ and PV-suitable area $A_{\mathrm{PV}}$ as well as the largest panel layout within the PV-suitable area are calculated.  
The method yields an average PV-suitable fraction of around $38$–$39\%$ of the predicted facade area after applying clearance buffers and component cleaning depending on the module size (Table \ref{tab:placement_loss}). Given the known facade and panel dimensions, the fitted layout was converted to installed peak capacity (kWp). Qualitative examples of the resulting layouts is presented in Figure~\ref{FIG:workflow_demo} and discussed further in Appendix~\ref{app:fail}, which illustrates common failure modes related to occlusions, segmentation accuracy, and facade complexity.

\textbf{Theoretical vs practical panel placement.} Figure~\ref{FIG:workflow_demo} shows the outputs of each step in the proposed workflow for eight example facades from different cities. Applying the panel layout generation procedure reveals cases where large PV-suitable areas in the binary mask do not translate into practical systems.  
In total, 14 facades with $f_{\mathrm{PV}} > 0.30$ yield a panel count of zero due to high ornamentation and narrow or irregular mask regions that cannot fit a single full-size \textit{S} module with clearances. The fraction of facade area that can be practically installed with solar panels (\textit{L–S} size) range between $4-5\%$, compared to a mean theoretical PV-suitable share of around $39\%$ across the LSAA subset (Table~\ref{tab:placement_loss}).  
The results confirm that relying on theoretical availability can overestimate installable capacity, particularly for highly articulated facades, as observed in the application dataset representing the old building stock in different cities. 

\begin{table}[t]
\centering
\small
\begin{tabular}{ccc}
\toprule

Wall surface (\%) & Theoretically suitable (\%) & Practically suitable (\%) \\
\midrule
54.01 & 38.62–39.48 & 4.35–4.75 \\

\bottomrule
\end{tabular}
\caption{Average shares of PV-suitable surfaces (\textit{large}–\textit{small} panels) for the LSAA subset.}
\label{tab:placement_loss}
\end{table}

\begin{figure}[!htbp]
	\centering
		\includegraphics[scale=.41]{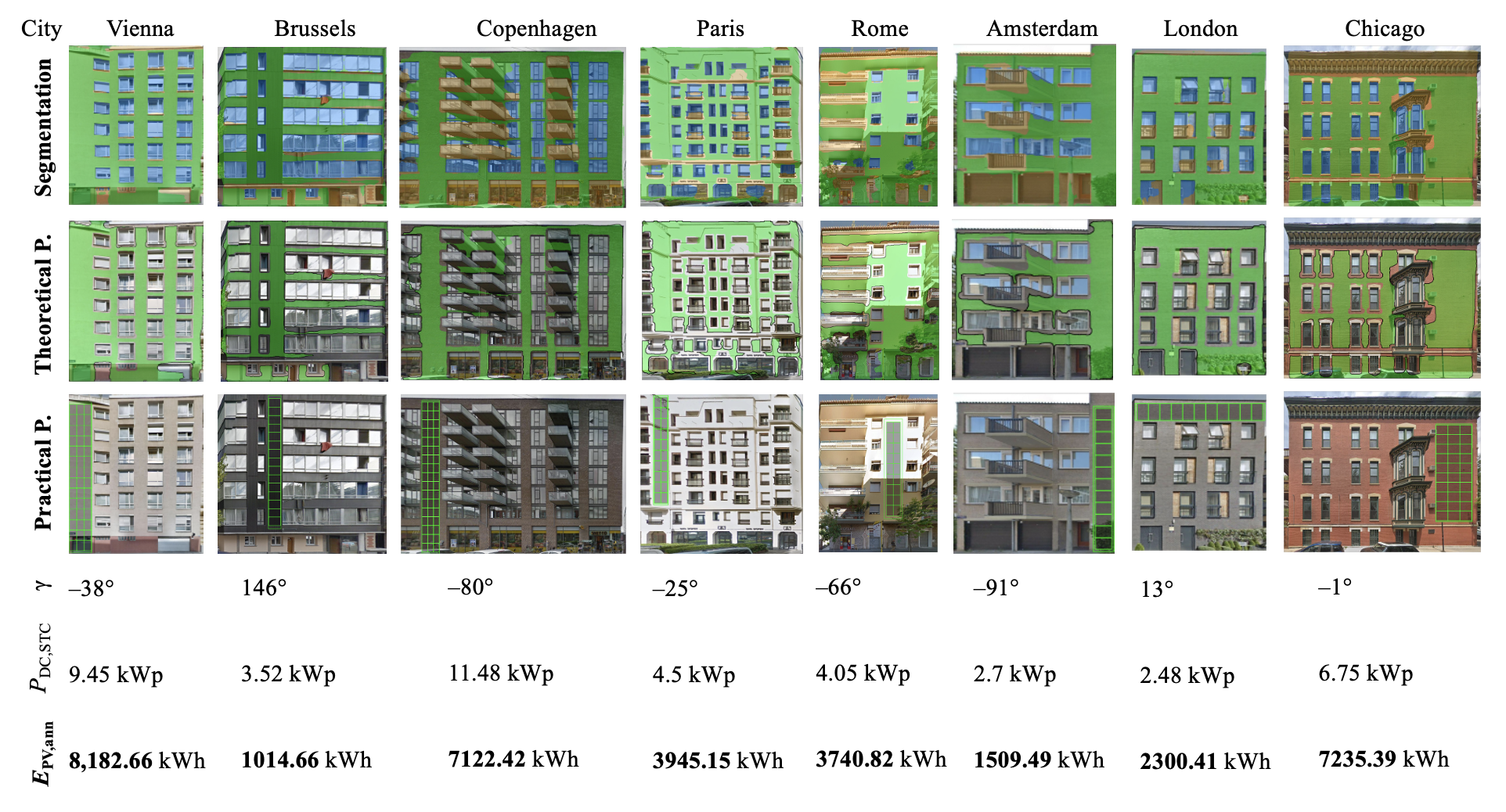}
    	\caption{Example applications (with panel size \textit{L}) from the LSAA subset (N = 373).}
	\label{FIG:workflow_demo}
\end{figure}

\section{Conclusion}

This study introduces a workflow that generates PV-suitability masks from facade images and translates them into practical panel layouts and energy estimates. By combining semantic segmentation and computational design methods, the workflow moves beyond theoretical potential estimation to produce layouts that can potentially be installed in practice. Evaluation on the CMP Facades dataset shows competitive performance across major facade classes, and application to a subset of the LSAA dataset demonstrates scalability across cities. By automatically estimating $A_{\mathrm{PV}}$ for large building stocks, the approach helps prioritize retrofits toward facades with large, uninterrupted wall fractions. When converted to kWp and paired with irradiance profiles, the results can support distribution-network planning by mapping potential generation in spatial and temporal dimensions.

The results for the PV layouts are based on a common \textit{L} and \textit{S} BIdPV panel sizes for facades. Varied or custom-sized panels could increase practical potential but would likely incur higher upfront costs. The generated PV layouts represent the maximum semantically suitable surfaces on facades within the given geometric constraints, excluding potential aesthetic considerations. Our method generates PV panel layouts using images and PVGIS integration to predict maximum energy yields by considering site and orientation specific energy yields. However, if 3D models are available, the accuracy of energy yield estimations can be improved through radiation simulations, particularly for surfaces that are substantially shaded. The results support early stage decision making rather than permitting decisions. Future work will focus on evaluating the generated PV layouts, improving the accuracy of segmentation performance, and generating design alternatives beyond the maximum panel coverage to help decision making in facade retrofits with BIPV.





\small
\bibliographystyle{unsrtnat}
\bibliography{references}

\newpage

\appendix
\section{Appendix}
\subsection{Metric Definitions} \label{app:metrics}

This appendix defines the evaluation metrics used in Section~\ref{sec:results}, which quantify different aspects of segmentation accuracy, class balance, and pixel-share estimation.  
Let $\mathrm{TP}_c$, $\mathrm{FP}_c$, and $\mathrm{FN}_c$ denote the true positive, false positive, and false negative pixel counts for class $c$, and let $C$ be the set of evaluated classes (all except \textit{unknown}).

Intersection-over-Union (IoU), measures the overlap between predicted and ground truth regions for class $c$, relative to their union:
\begin{equation}
\mathrm{IoU}_c = \frac{\mathrm{TP}_c}{\mathrm{TP}_c + \mathrm{FP}_c + \mathrm{FN}_c}.
\end{equation}

Mean Intersection-over-Union (mIoU), summarizes overall segmentation performance across all classes:
\begin{equation}
\mathrm{mIoU} = \frac{1}{|C|} \sum_{c \in C} \mathrm{IoU}_c.
\end{equation}
Two variants are reported: (1) dataset-level mIoU, computed from aggregated counts across all test images; and (2) mean over per-image mIoU values.

Macro Precision and Recall, indicate, respectively, the proportion of predicted pixels that are correct and the proportion of ground truth pixels that are recovered, averaged equally over classes. Macro recall corresponds to “Average class accuracy” reported in \cite{gadde_efficient_2016}.:
\begin{align}
\mathrm{Precision} &= \frac{1}{|C|} \sum_{c \in C} \frac{\mathrm{TP}_c}{\mathrm{TP}_c + \mathrm{FP}_c}, \\
\mathrm{Recall} &= \frac{1}{|C|} \sum_{c \in C} \frac{\mathrm{TP}_c}{\mathrm{TP}_c + \mathrm{FN}_c}.
\end{align}

Macro F1 score is the harmonic mean of macro precision and recall, giving a single measure of class-balanced accuracy:
\begin{equation}
\mathrm{F1} = \frac{2 \cdot \mathrm{Precision} \cdot \mathrm{Recall}}{\mathrm{Precision} + \mathrm{Recall}}.
\end{equation}

Pixel Accuracy (Overall) is the proportion of correctly classified pixels over all non-ignored pixels:
\[
\mathrm{Acc} = \frac{\sum_{c \in \tilde{C}} \mathrm{TP}_c}{\sum_{c \in \tilde{C}} (\mathrm{TP}_c + \mathrm{FN}_c)}
\]
where $\tilde{C}$ denotes all non-ignored labels (including class 0). 
This corresponds to “Overall” in \cite{gadde_efficient_2016}.

Class-share consistency evaluates how well the predicted pixel share of each class matches the ground truth.
For image $i$, let $s_{c,i}$ be the predicted share of pixels for class $c$ and $s^\mathrm{GT}_{c,i}$ the ground-truth share. The signed error for class $c$ is
\[
\Delta s_{c,i} = s_{c,i} - s^\mathrm{GT}_{c,i} \quad (\mathrm{pp}),
\]
where ``pp'' denotes percentage points.  
The mean absolute error (MAE) and root-mean-square error (RMSE) across $N$ images are:
\begin{align}
\mathrm{MAE}_c &= \frac{1}{N} \sum_{i=1}^N |\Delta s_{c,i}|, \\
\mathrm{RMSE}_c &= \sqrt{\frac{1}{N} \sum_{i=1}^N \left(\Delta s_{c,i}\right)^2}.
\end{align}

\subsection{Data Augmentation Details}\label{app:aug}
All enhancements are applied on the fly during training at a resolution of $640 \times 640$. Transformations include horizontal flips with probability $p = 0.5$; random shifts of up to $\pm 10\%$, scaling between $0.9$ and $1.1$, and rotations of up to $\pm 5^\circ$; brightness and contrast adjustments with factors in $[0.9, 1.1]$; HSV jitter with hue shifts of $\pm 10^\circ$ and saturation and value scaling in $[0.9, 1.1]$; and Gaussian blur with a kernel of $3 \times 3$ and $\sigma \in [0.0, 1.0]$.

\subsection{Hyperparameter Tuning}\label{app:hyper}
To set optimizer and training hyperparameters, we carried out a small grid search on the CMP validation split. For each configuration, we trained the SegFormer model initialized from ADE20K for up to 80 epochs with a cosine learning-rate schedule and linear warmup (warmup ratio $0.05$), using early stopping when the validation mIoU did not improve for 10 consecutive epochs. The search space comprised learning rates $\{5{\times}10^{-5},\,10^{-4},\,2{\times}10^{-4}\}$, batch sizes $\{2,4\}$, and gradient accumulation steps $\{1,2\}$, with weight decay fixed at $10^{-2}$. The final model uses the configuration that achieved the highest validation mIoU under the sweep: learning rate $2{\times}10^{-4}$, weight decay $10^{-2}$, warmup ratio $0.05$, batch size $4$, and accumulation steps $1$.

\subsection{Panel Placement Parameters} \label{app:panel_placement}
The placement of modules is performed for \textit{large} ($935 \times 1300$\,mm) and \textit{small} ($720 \times 875$\,mm) facade modules\footnote{\label{3s}\url{https://www.3s-solar.swiss/downloads-teraslate\#systemeproduktes}} with a fixed gap between panels of $0.02$\,m. Both portrait and landscape orientations are tested.  

The binary PV mask has image dimensions $(W_{\mathrm{px}}, H_{\mathrm{px}})$ and known facade size $(W_{\mathrm{m}}, H_{\mathrm{m}})$. The scale factors are
\[
s_x = \tfrac{W_{\mathrm{px}}}{W_{\mathrm{m}}}, \quad 
s_y = \tfrac{H_{\mathrm{px}}}{H_{\mathrm{m}}},
\]
allowing panel dimensions and gaps to be expressed in pixels.  

To locate the largest axis-aligned rectangle within the mask, we build row-wise histograms of consecutive ``PV'' pixels per column and compute the maximal rectangle per histogram using a standard stack-based algorithm. This ensures the rectangle is fully contained in the PV region.  

Within the rectangle, the number of modules per axis is computed as
\[
n_x = \Big\lfloor \frac{R_w + g}{p_w + g} \Big\rfloor, \quad
n_y = \Big\lfloor \frac{R_h + g}{p_h + g} \Big\rfloor,
\]
where $(R_w,R_h)$ are the rectangle dimensions in pixels, $(p_w,p_h)$ the panel size in pixels, and $g$ the inter-panel gap in pixels. Panels are accepted only if fully contained; partial panels are excluded. The resulting grid is centered in the rectangle.

\subsection{Energy Yield Estimation} \label{app:pvgis}


The annual energy yield is computed from the installed peak DC capacity depending on the module type (\textit{large} $= 225$ Wp, \textit{small} $= 110$ Wp)\textsuperscript{\ref{3s}} and the PVGIS-returned annual specific yield:
\begin{equation}
E_{\mathrm{PV,ann}} \;=\; P_{\mathrm{DC,STC}} \times Y_{\mathrm{spec}}
\quad (\mathrm{kWh/yr}),
\label{eq:epv_main}
\end{equation}
where \(P_{\mathrm{DC,STC}}\) is in \(\mathrm{W_p}\) (from \autoref{panel_layout_energy}) and \(Y_{\mathrm{spec}}\) is the location- and orientation-specific annual specific yield \([\mathrm{kWh/kWp/yr}]\) returned by PVGIS given the inputs (tilt \(90^\circ\), facade azimuth, integrated mounting, and default loss parameters: cable 1\%, inverter 2\%, PV 0.5\%).

\begin{equation}
Y_{\mathrm{spec}} \;=\; \mathrm{PR}_{\mathrm{ann}} \;\times\; \frac{H_{\mathrm{POA,ann}}}{G_{\mathrm{STC}}},
\qquad
G_{\mathrm{STC}} = 1~\mathrm{kW/m^2},
\label{eq:yspec_decomp}
\end{equation}
with \(H_{\mathrm{POA,ann}}\) the annual plane-of-array irradiation \([\mathrm{kWh/m^2/yr}]\) and \(\mathrm{PR}_{\mathrm{ann}}\) the annual performance ratio (\%) accounting for temperature, spectral and angle-of-incidence effects, and system losses as modeled by PVGIS. Substituting \autoref{eq:yspec_decomp} into \autoref{eq:epv_main} yields:
\begin{equation}
E_{\mathrm{PV,ann}} = P_{\mathrm{DC,STC}} \,
\mathrm{PR}_{\mathrm{ann}} \,
\frac{H_{\mathrm{POA,ann}}}{G_{\mathrm{STC}}},
\end{equation}

\subsection{Common Failure Modes in Layout Generation} \label{app:fail}

\begin{figure}[h!]
	\centering
		\includegraphics[scale=.36]{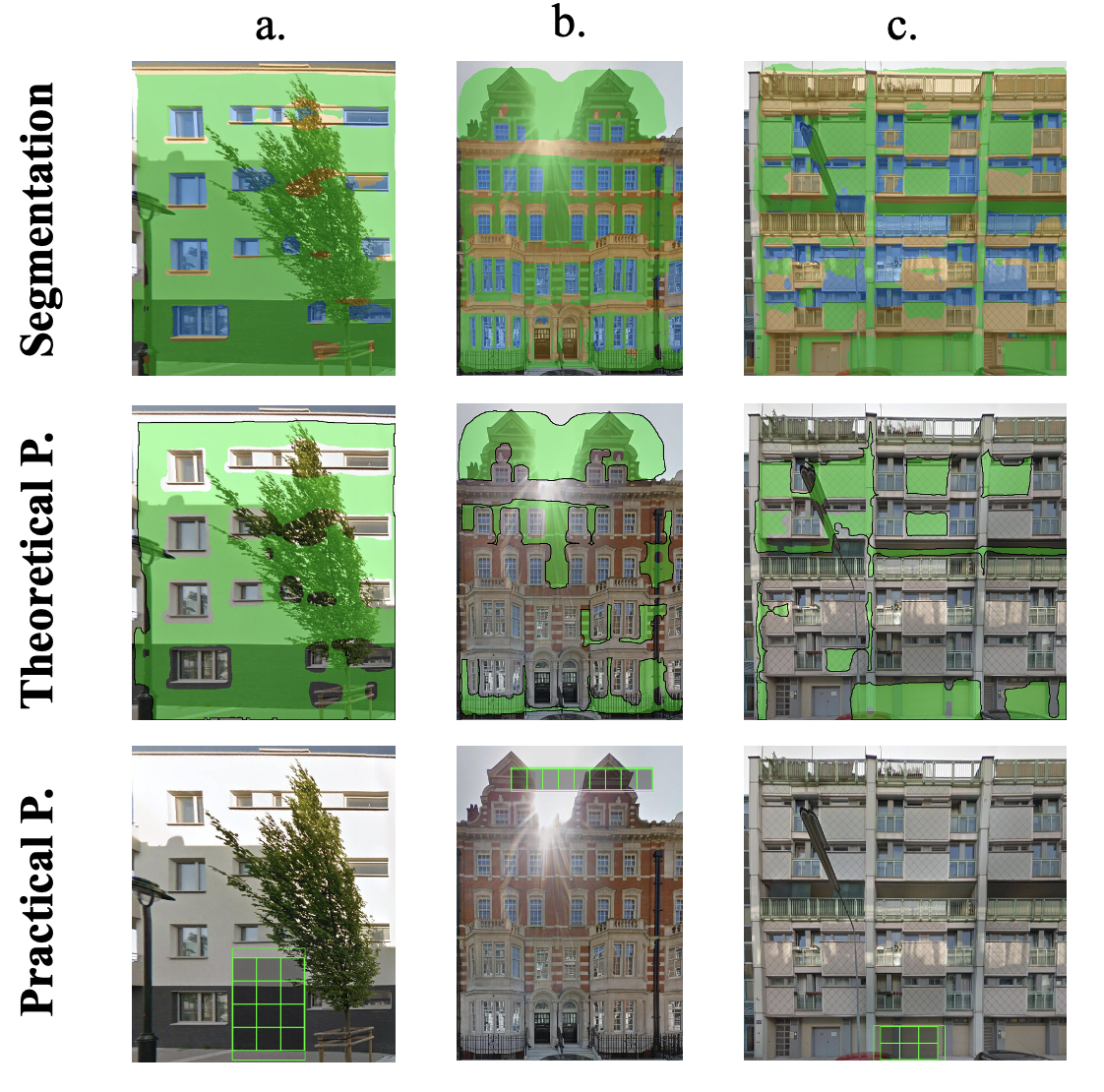}
    	\caption{Failure modes due to occlusions (a), segmentation accuracy (b), complexity of facade (c).}
        \vspace*{9in}
	\label{FIG:fail}
\end{figure}

A qualitative assessment shows three recurring failure modes (Fig.\ref{FIG:fail}a–c). Foreground occlusions, such as trees, people, vehicles, trees, are not explicitly segmented and are thus treated as walls, resulting in unreliable PV areas (Fig.\ref{FIG:fail}a). This can be mitigated by an occlusion-removal step, such as detecting and masking foreground objects or inpainting, and by adding an explicit \emph{occlusion/vegetation} class, especially for trees that may cast shade. Pitched roofs are another source of error due to limited representation in CMP, leading to poor segmentation on sloped geometries (Fig.\ref{FIG:fail}b). Adding pitched-roof examples and marking such these as non-suitable can improve robustness. Finally, highly articulated or repetitive facades, such as textured surfaces, and arrays of balconies, reduce segmentation quality and expose a limitation of the largest-rectangle heuristic, which overlooks multiple smaller feasible regions (Fig.\ref{FIG:fail}c). Allowing multi-region packing, in other words placing panels across several disjoint areas, relaxing the single-rectangle assumption, and enforcing minimum cluster sizes can yield layouts with larger energy generation capacity.

\end{document}